%% file: main.tex
\documentclass[10pt,twocolumn,letterpaper]{article}

\usepackage[pagenumbers]{cvpr} 

\input{preamble}

\definecolor{cvprblue}{rgb}{0.21,0.49,0.74}
\usepackage[pagebackref,breaklinks,colorlinks,citecolor=cvprblue]{hyperref}
\usepackage{colortbl}

\usepackage{savesym}
\savesymbol{todo}
\usepackage[colorinlistoftodos,prependcaption,textsize=small]{todonotes}
\usepackage{xargs}                      
\usepackage{wrapfig}
\usepackage{listings}
\makeatletter
\robustify\@latex@warning@no@line
\makeatother
\usepackage{authblk}
\newcommand\figcaption{\def\@captype{figure}\caption}
\newcommand\tabcaption{\def\@captype{table}\caption}\makeatother

\title{Training Matting Models without Alpha Labels}

\author[1]{Wenze Liu}
\author[2]{Zixuan Ye}
\author[2]{Hao Lu$^\ast$}
\author[2]{Zhiguo Cao}
\author[1]{Xiangyu Yue\thanks{Corresponding authors}}
\affil[1]{Multimedia Lab, The Chinese University of Hong Kong}
\affil[2]{School of Artificial Intelligence and Automation, Huazhong University of Science and Technology}

\begin{document}
\maketitle
\input{sec/0_abstract}    
\input{sec/1_intro}
\input{sec/2_relatedwork}
\input{sec/3_method}
\input{sec/4_experiment}
\input{sec/5_conclusion}

{
    \small
    \bibliographystyle{ieeenat_fullname}
    \bibliography{main}
}

\input{sec/X_suppl}

\end{document}

%% file: preamble.tex
\usepackage[dvipsnames]{xcolor}


%% file: sec/0_abstract.tex
\begin{abstract}
The labelling difficulty has been a longstanding problem in deep image matting. To escape from fine labels, this work explores using rough annotations such as trimaps coarsely indicating the foreground/background as supervision. We present that the cooperation between learned semantics from indicated known regions and proper assumed matting rules can help infer alpha values at transition areas. Inspired by the nonlocal principle in traditional image matting, we build a directional distance consistency loss (DDC loss) at each pixel neighborhood to constrain the alpha values conditioned on the input image. DDC loss forces the distance of similar pairs on the alpha matte and on its corresponding image to be consistent. In this way, the alpha values can be propagated from learned known regions to unknown transition areas. With only images and trimaps, a matting model can be trained under the supervision of a known loss and the proposed DDC loss. Experiments on AM-2K and P3M-10K dataset show that our paradigm achieves comparable performance with the fine-label-supervised baseline, while sometimes offers even more satisfying results than human-labelled ground truth. Code is available at \url{https://github.com/poppuppy/alpha-free-matting}.
\end{abstract}

%% file: sec/1_intro.tex
\section{Introduction}
\label{sec:intro}
Image matting, a fundamental task in image editing, aims to decompose an input image $\boldsymbol{I}$ into two layers, \ie, the foreground $\boldsymbol{F}$ and the background $\boldsymbol{B}$. Specifically, it estimates the foreground opacity $\boldsymbol{\alpha}$ a.k.a. alpha matte as 
\begin{equation}
  \boldsymbol{I} = \boldsymbol{\alpha} \boldsymbol{F} + (1 - \boldsymbol{\alpha}) \boldsymbol{B}\,.
  \label{eq:matting}
\end{equation}
Recent years have witnessed dramatic improvement in image matting techniques~\cite{hou2019context,li2020natural,liu2021long,liu2021tripartite,tang2019learning,park2022matteformer,yao2023vitmatte,yu2021mask}, particularly since Deep Image Matting (DIM)~\cite{xu2017deep} pioneered an end-to-end training paradigm. Despite the prosperity of the labelling-and-training paradigm, large-scale datasets are difficult to collect for the subtle details required. Existing datasets either composite~\cite{xu2017deep} few annotated foreground objects with thousands of backgrounds, or make small special-purpose~\cite{li2022bridging,li2021privacy} datasets. Although the deficiency of datasets is regarded as a bottleneck of deep image matting, the contradiction between the need of large-scale datasets and the labeling difficulty has yet to be handled.
\begin{figure}[!t]
	\centering
	\includegraphics[width=\linewidth]{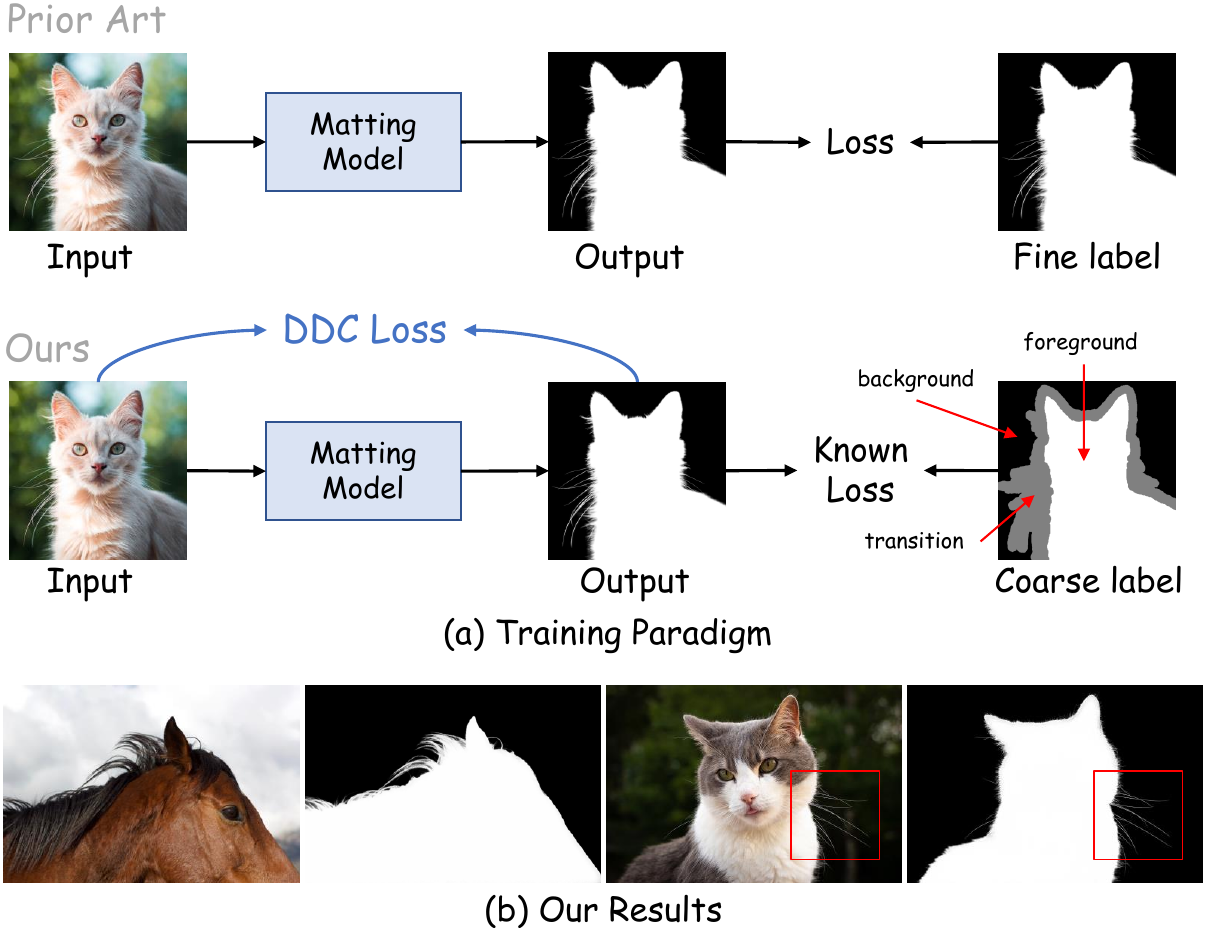}
	\caption{\textbf{Our image matting training paradigm without alpha labels.} (a) Compared with prior art adopting fine alpha matte annotations, we use only coarse trimaps as labels. During the training phase, we use an $l_1$ loss termed known loss to supervise the known regions indicated by the trimap, and devise a DDC loss to restrict the alpha values at transition areas. (b) Even without fine annotated labels, our trained model predicts accurate alpha mattes.
	}
	\label{fig:paradigm}
\end{figure}

Solving $\boldsymbol{\alpha}$ from Eq.~\eqref{eq:matting} has not always been relying on labels, though. Aided by user prompts, \eg, trimaps indicating the known ($1$ for foreground and $0$ for background), and the unknown ($0.5$ for transition) regions as shown in Fig.~\ref{fig:paradigm}, many works~\cite{chuang2001bayesian,sun2004poisson,levin2007closed,he2010fast,lee2011nonlocal,chen2013knn} try to trace the color cues to infer alpha values from known to unknown. For example, Closed Form Matting~\cite{levin2007closed} assumes the color line model in each neighborhood to figure out the alpha matte. Nonlocal Matting~\cite{lee2011nonlocal} borrows the nonlocal principle from image denoising~\cite{buades2005review,buades2008nonlocal} to solve the alpha value with affinity matrices. These traditional methods tend to be impaired in the face of complex textures or similar colors on the junction of foreground and background~\cite{xu2017deep}, because they primarily consider low-level color or texture information. In contrast, deep learning based methods inject semantic understanding to adapt the models to complex scenarios, resulting in a significant performance improvement. Along the torrent, recent studies~\cite{li2021deep,li2022bridging,qiao2020attention,zhang2019late,li2021privacy,wei2021improved,yang2022unified} prove that strong semantics introduced by deep models can reduce the usage costs, either by simplifying user prompts~\cite{wei2021improved,yang2022unified} or exploring automatic image matting~\cite{li2021deep,li2022bridging,qiao2020attention,zhang2019late,li2021privacy} with no user prompt required. Different from research efforts devoted on the user end, we would like to study whether a collaborative approach involving robust semantic understanding and well-defined matting rules can mitigate the need for fine annotations during the training phase. 

In this work, we demonstrate that, with proper supervision, deep image matting models can be effectively trained without any fine annotations. For one thing, deep models are validated~\cite{long2015fully,xiao2018unified,kirillov2019panoptic} to be expert at learning the semantics from coarse labels. For another, traditional solutions~\cite{levin2007closed,chen2013knn} have proven that proper assumed rules can propagate the matting constraints from known to unknown regions. Based on these existing experiences, we conjecture that the labels only need to provide rough semantics, while not necessary to indicate the transparency at transition areas. We thus build a preliminary matting supervision, where an $l_1$ loss termed known loss supervises the known areas indicated by the trimap, and the unknown ones are supervised using the nonlocal principle. The preliminary experiments suggests that i) the deep model can utilize the information learned from known regions to infer unknown ones, but in the sense of hard segmentation, and ii) the additional nonlocal assumption helps produce some details at transition areas. Based on further observation and analysis, we find that the embodiment of nonlocal principle in~\cite{lee2011nonlocal,chen2013knn} is unsuited for the deep learning process, due to the `braking effect' and `hard segmentation effect'. The former impedes the refinement of details with long-range dependency, while the latter hurts smooth transition of alpha values at boundaries. Aiming at the two problems, we introduce a novel expression of nonlocal principle as the loss function, called \textit{distance consistency loss} (DC loss). DC loss forces the euclidean distance between each pixel and several similar neighbors in the predicted alpha matte to be equal with that in the image. DC loss well addresses the two issues above, but introduces undesired texture noise in interior regions, where it is incompatible with known loss. With a detailed analysis, we update DC loss with \textit{directional distance consistency loss} (DDC loss), which solves the conflict between the two losses and eliminates the interior noise with no side effects.

In summary, we establish a novel training paradigm for image matting based on the proposed DDC loss as shown in Fig.~\ref{fig:paradigm}. With only trimaps as labels, the model trained under our proposed paradigm can predict fine details, \eg, the cat beard as shown on the right of Fig.~\ref{fig:paradigm}. Experiment results on animal~\cite{li2022bridging} and portrait~\cite{li2021privacy} matting datasets show that our models performs comparably with the baseline supervised by fine alpha labels, which verifies the feasibility and effectiveness of the proposed paradigm. Further more, we provide detailed illustration and analysis to shed light on the working principle of our approach.

%% file: sec/2_relatedwork.tex
\section{Related Work}
\label{sec:relatedwork}
\vspace{5pt}
\noindent\textbf{Traditional Image Matting}~\cite{chuang2001bayesian,levin2007closed,lee2011nonlocal,he2010fast,chen2013knn,aksoy2017designing} studies how to add appropriate prior knowledge reflecting the related physical laws into Eq.~\eqref{eq:matting}, to propagate the user specified constraints towards the alpha matte solution. Bayesian Matting~\cite{chuang2001bayesian} adopts Bayesian estimation to infer the unknown alpha values based on color sampling. Closed Form Matting~\cite{levin2007closed} assumes certain smoothness, \ie, the color line model, of the foreground and background in a small neighborhood, based on which a matting Laplacian matrix is derived. It yields impressive results when the color line model assumption holds. In order to relax the conditions, Nonlocal Matting~\cite{lee2011nonlocal} introduces nonlocal principle, a conception originally adopted in image denosing~\cite{buades2005review,buades2008nonlocal} tasks. Nonlocal Matting expresses the nonlocal principle via an affinity matrix to form a constraint condition, which is then combined with the user constraint to solve the alpha matte. Follow up work such as Fast Matting~\cite{he2010fast} and KNN Matting~\cite{chen2013knn} studies the construction of the affinity matrix. These traditional methods consider only the low-level color and texture information, while incapable of semantic perception. Hence, they struggle to work in complex scenarios, \eg, where the foreground and the background have similar colors. Our work takes in matting priors, but for building a feasible training mode without alpha labels in the deep matting era.
\begin{figure*}[!t]
	\centering
	\includegraphics[width=\linewidth]{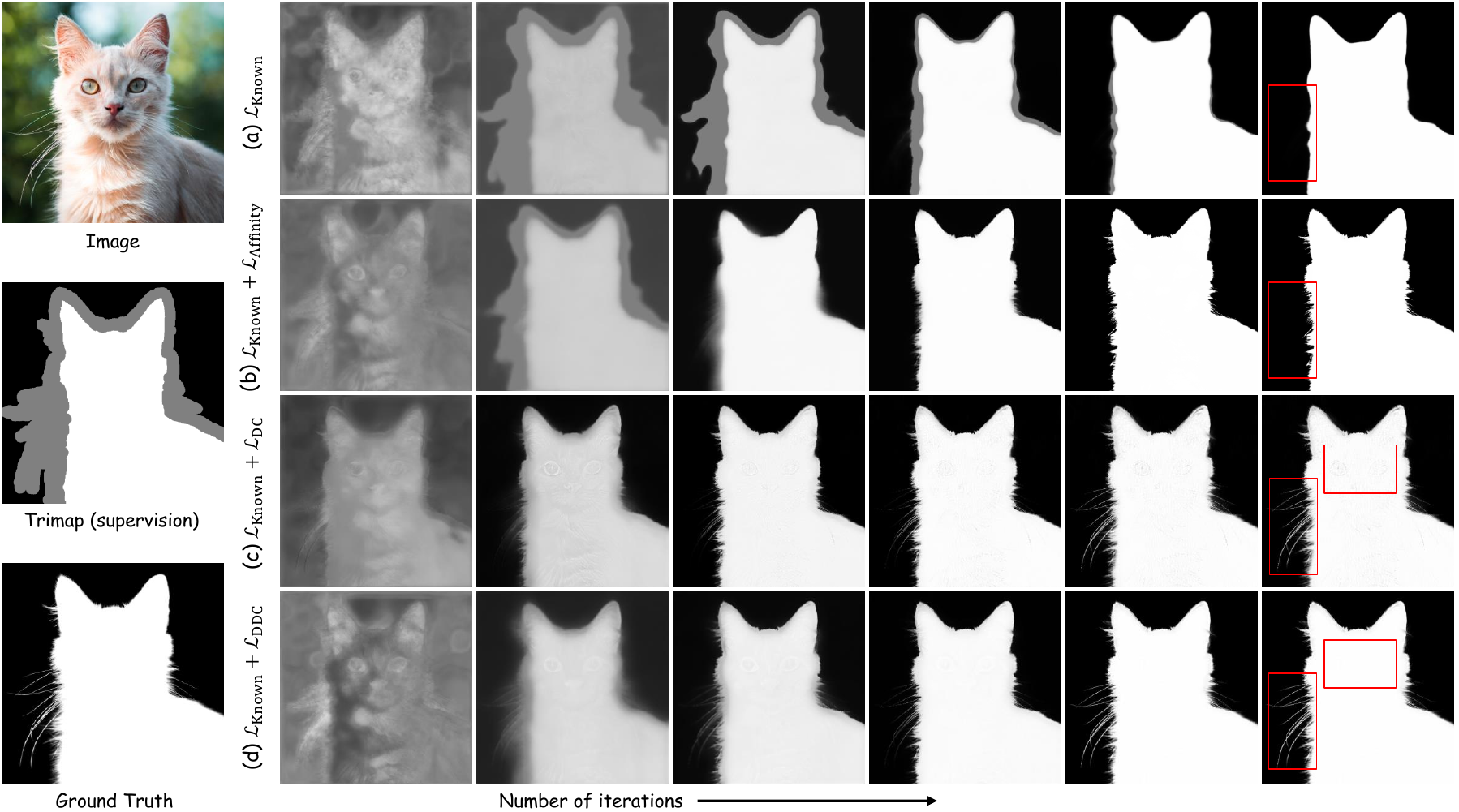}
	\caption{\textbf{The outputs during training of four supervision policies.} (a) The model can learn to extend the semantics from unknown to known with known loss. (b) The cooperation between known loss and affinity loss helps predict details, but fails to delineate long hair and causes hard segmentation. (c) The proposed DC loss well fits long hair and smooth transition on boundaries, but introduces texture noise on the foreground.
    (d) DDC loss eliminates the interior noise with no side effects. 
	}
	\label{fig:traing_alpha}
\end{figure*}

\noindent\textbf{Deep Image Matting}, first presented in~\cite{shen2016deep,cho2016natural,xu2017deep}, has opened a new era for modern image matting methods. A highly-performed deep matting model can be trained with a ready-made dataset providing the inputs and labels. The label is the manually annotated alpha matte, while the input varies over the development of deep matting techniques. Early work~\cite{lu2019indices,hou2019context,li2020natural,liu2021long,liu2021tripartite,tang2019learning,park2022matteformer,yao2023vitmatte,yu2021mask,cai2022transmatting} takes both the image and a trimap indicating the smooth/transition areas as the input, and mainly studies the network architecture to better adapt the high-level semantics and low-level details for the matting task. In the interests of reducing the user efforts, there recently appears some attempts that replace the auxiliary trimap input with simpler prompts such as scribble~\cite{xiao2018unified} and click~\cite{wei2021improved}. Another branch of work~\cite{li2021deep,li2022bridging,qiao2020attention,zhang2019late,li2021privacy,ma2023rethinking,ke2022modnet,chen2018semantic} studies to fully get rid of auxiliary inputs, leading to an automatic matting manner, where the model only takes in the image and needs to find salient objects itself. Besides simplifying the inputs from the user side, several work notice the deficiency of datasets. Despite the enrichment of capacity and variety by some datasets, they still faces problems of unrealistic contents~\cite{xu2017deep} and few categories~\cite{li2022bridging,li2021privacy}. There is also attempt~\cite{liu2020boosting} to add coarse training labels to lift the matting performance. With the same spirit of alleviating the data deficiency, we instead set out from lightening of annotation burdens--simplifying the labels. Rather than using the fine alpha matte, we explore to supervise the model with coarse trimaps, and employ our paradigm on automatic matting as application examples.

%% file: sec/3_method.tex
\section{Method}
In Sec.~\ref{subsec:preliminary}, we first provide a preliminary study on the possibility of using trimaps as labels during training deep matting models. In what follows, we analyse the deficiency of the form of nonlocal principle in prior art for deep learning process. In Sec.~\ref{subsec:dcloss}, we present our novel expression of nonlocal principle, and introduce our DC loss. At last in Sec.~\ref{subsec:ddcloss}, we solve the conflict problem between known loss and DC loss by updating DC loss to DDC loss.
\label{sec:method}
\subsection{Exploration and Analysis}
\label{subsec:preliminary}
We adopt the recent image matting model ViTMatte-S~\cite{yao2023vitmatte} and the AM-2K dataset~\cite{li2022bridging} for preliminary experiments. The detailed training configurations can be found in the experiment section.

\vspace{5pt}
\noindent\textbf{Learning semantics from trimaps.} We set the image as input and the corresponding trimap as training label respectively. An $l_1$ loss termed known loss is set to supervise only the known regions (indicated by $0$ and $1$ in trimaps). An instance on the evolution process of the output during training is shown in Fig.~\ref{fig:traing_alpha} (a). The model quickly finds the foreground object in early stage, then learns to fit the shape of the trimap, and finally the unknown region shrinks to disappearance. This phenomenon explains that the learned semantics can be extended to unsupervised regions. However, known loss can only provide low-quality segmentation results, and can not generate details.

\begin{figure}[!t]
	\centering
	\includegraphics[width=\linewidth]{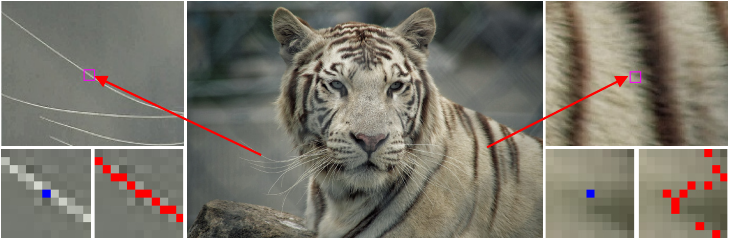}
	\caption{\textbf{Finding similar pixels in a local window.} Centered at a pixel (blue) in the image, the top $K$ similar pixels (red) are selected in each $K\times K$ local window (pink) according to the euclidean distance.
	}
	\label{fig:topk}
\end{figure}

\vspace{5pt}
\noindent\textbf{Learning alpha values at unknown regions.} In the second experiment, we try to add constraints with nonlocal principle~\cite{lee2011nonlocal}. In image matting, the nonlocal principle claims that the alpha value $\alpha_i$ of a pixel $\boldsymbol{I}_i$ is a weighted sum of alpha values of pixels similar to $\boldsymbol{I}_i$, formulated by
\begin{equation}
  \boldsymbol{A}\boldsymbol{\alpha}=\boldsymbol{\alpha}\,.
  \label{eq:nonlocal}
\end{equation}
Note that $\boldsymbol{\alpha}$ is flattened to $N\times1$, where $N$ indicates the number of pixels, and $\boldsymbol{A}$ is the affinity matrix of size $N\times N$. In KNN Matting~\cite{chen2013knn}, $\boldsymbol{A}$ is calculated as 
\begin{equation}
  \boldsymbol{A}(i,j)=1-\frac{\Vert \boldsymbol{I}_i - \boldsymbol{I}_j\Vert_2}{C}\,,j\in \textrm{argtopk}\{-f(i,j)\}\,,
  \label{eq:affinity}
\end{equation}
and then row-normalized, where $C$ is a constant to make $\boldsymbol{A}(i,j)\in[0,1]$, $f$ is set as $f(i,j)=\Vert \boldsymbol{I}_i - \boldsymbol{I}_j\Vert_2 + d_{ij}^2$, and $d_{ij}$ denotes the spatial distance between $i$ and $j$. Though Eq.~\eqref{eq:affinity} takes the whole image into account, the item $d_{ij}$ limits the high response pixels within a small neighborhood for each pixel. Meanwhile, $d_{ij}$ introduces noise into the similarity calculation of the pixel values (see also supplementary materials). In order to reduce both the computational workload and the noise introduced, we choose to calculate the affinity matrix $\boldsymbol{A}$ within a fixed $K\times K$ window centered at each pixel, where $K\geq3$ is an odd number. Specifically, $\boldsymbol{A}$ is initialized as all zeros, and we calculate the similarity score between the central pixel and all pixels in the window with $f(i,j)=\Vert \boldsymbol{I}_i - \boldsymbol{I}_j\Vert_2$, and fill only top $K$ scores into $\boldsymbol{A}$ according to the indices. An instance for similar pixel selection is exhibited in Fig.~\ref{fig:topk}. On the basis of supervising the known regions with $l_1$ loss, the affinity loss
\begin{equation}
  \mathcal{L}_\textrm{affinity}=\frac{1}{N}\Vert \boldsymbol{A}\boldsymbol{\alpha}-\boldsymbol{\alpha}\Vert_1\,
  \label{eq:affinity_loss}
\end{equation}
is added onto the whole alpha matte. As shown in Fig.~\ref{fig:traing_alpha} (b), the added loss invites subtle details to the alpha matte. This verifies the possibility of using only trimaps as labels for training matting models.

Though affinity loss $\mathcal{L}_\textrm{affinity}$ helps predict subtle details, it encounters two problems: i) it cannot predict details with long-range dependency, \eg, long hair, and ii) it does dot encourage smooth transition of alpha values at boundaries. The problems will be elaborated as follows.
\begin{figure}[!t]
	\centering
	\includegraphics[width=\linewidth]{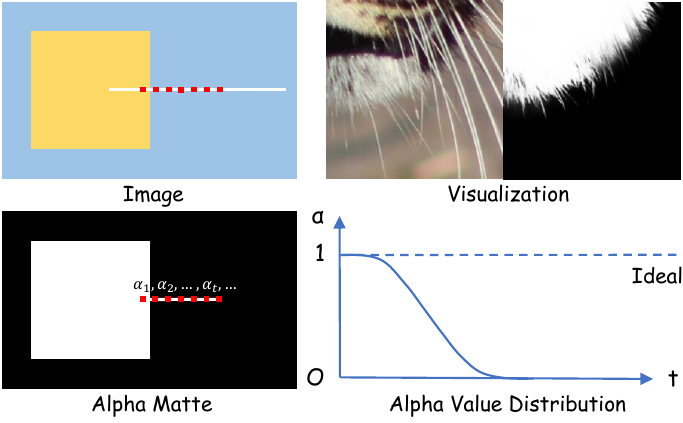}
	\caption{\textbf{The growth arrest problem of long hair supervised by affinity loss.} Suppose there is a yellow tiger with a hair whose intensity is a constant in the image. We derive that under such conditions alpha values of approximate linear or quadratic variation produce small loss. Considering that the already generated coarse mask provides the initial starting and ending value as $1$ and $0$, the distribution of alpha values are stuck at a state shown in the bottom right plot.
	}
	\label{fig:braking}
\end{figure}

\vspace{5pt}
\noindent\textbf{Braking effect.}
We have a close look at the failure of predicting details with long-range dependency supervised by affinity loss. Fig.~\ref{fig:braking} provides a visualization and a simulated example on long hair prediction. Let the blue color denote the background, and the yellow be a tiger with a hair of one pixel wide. For convenience we assume that the hair color is homogeneous, so that given a pixel on the hair, the selected similar pixels evenly distribute in both sides and the normalized weights are all $\frac{1}{K}$ in $\boldsymbol{A}$. Since the model first generates coarse mask for the foreground object and then starts to delineate the fine details (cf. Fig.~\ref{fig:traing_alpha} (b), from left to right), here we suppose the tiger body has been correctly predicted, and use a sequence $\{\alpha_t\}$ to denote the alpha values corresponding to the hair, from left to right. Then the condition $\boldsymbol{A}\boldsymbol{\alpha}=\boldsymbol{\alpha}$ is equivalent to 
\begin{equation}
  \frac{1}{K}(S_{t-1}-S_{t-K}+\alpha_t)=\alpha_{t-\frac{K-1}{2}}\,,
  \label{eq:sequence_relation}
\end{equation}
where $S_t=\sum\limits_{i=1}^t\alpha_i$. Then $\{\alpha_t\}$ has a recursion 
\begin{equation}
  \alpha_t=\alpha_{t-K}+K(\alpha_{t-\frac{K-1}{2}}-\alpha_{t-\frac{K+1}{2}})\,.
  \label{eq:sequence_recursion}
\end{equation}
Its characteristic equation $x^K-K(x^{\frac{K+1}{2}}-x^{\frac{K-1}{2}})-1=(x-1)^3(\frac{K^2-1}{8}x^{\frac{K-3}{2}}+\sum\limits_{i=1}^{\frac{K-3}{2}}\frac{i(i+1)}{2}(x^{i-1}+x^{K-2-i}))=0$ has three multiple real roots $1$ and $K-3$ complex roots. Given the monotonically non-increasing feature, $\{\alpha_t\}$ is a form of $\alpha_t=C_1+C_2t+C_3t^2$, where $C_1$, $C_2$, $C_3$ are constants. Besides being always $1$ as expected, $\{\alpha_t\}$ can also be approximate linear or quadratic variation over $t$ to produce small loss. Meanwhile, under the power of the known loss, the starting and ending value are initialized as $1$ and $0$ respectively. Under their influence $\{\alpha_t\}$ tends to distribute as in the bottom right subplot of Fig.~\ref{fig:braking}, implying that the hair growth gets stuck early. Enlarging the window size $K$ helps a little when the hair is similar to the body in color, because more body values are taken into account in windows responsible for calculating the first few $\alpha_t$'s. However, it is unhelpful for isolated hair according to the reasoning of Fig.~\ref{fig:braking}. Therefore the condition $\mathcal{L}_\textrm{affinity}=0$ does not satisfy our optimization objective.

\vspace{5pt}
\noindent\textbf{Hard segmentation effect.} 
$\mathcal{L}_\textrm{affinity}$ does not ensure smooth transitions at boundaries. Consider a certain part on the edge where the color change obeys linear variation horizontally in the image. Then its cross section can be described as $\boldsymbol{I}_i=\boldsymbol{a}x_i+\boldsymbol{b}$, where $x_i$ is the horizontal coordinate of pixel $i$, and $\boldsymbol{a}$, $\boldsymbol{b}$ are constant vectors. Given a pixel on the hair, the top similar pixels are evenly distributed on both sides, whose similarity scores can be calculated as 
\begin{equation}
  \boldsymbol{A}(i,j)=1-\frac{\Vert \boldsymbol{I}_i - \boldsymbol{I}_j\Vert_2}{C}=1-\frac{\vert x_i-x_j\vert\Vert \boldsymbol{a} \Vert_2}{C}\,.
  \label{eq:symmetrical_score}
\end{equation}
Therefore symmetrical pixels correspond to equal similarity scores. Let $w_i, i=1,2,...,K$ denote the row-normalized weights, where $w_i=w_{K-i}$. Now $\boldsymbol{A}\boldsymbol{\alpha}=\boldsymbol{\alpha}$ becomes $\sum\limits_{j=1}^K w_j\alpha_j=\alpha_i$. Then multiple situations satisfy $\boldsymbol{A}\boldsymbol{\alpha}=\boldsymbol{\alpha}$, \eg, $\alpha_i=mx_i+n$, where $m$ is an arbitrary number and $n$ can be chosen to meet other constraints. Because known loss encourages hard segmentation, the linear variation is squeezed to be more sharp similar to the bottom right subplot in Fig.~\ref{fig:braking}. In other words, the combination of the two losses contributes to hard segmentation.

\begin{figure}[!t]
	\centering
	\includegraphics[width=\linewidth]{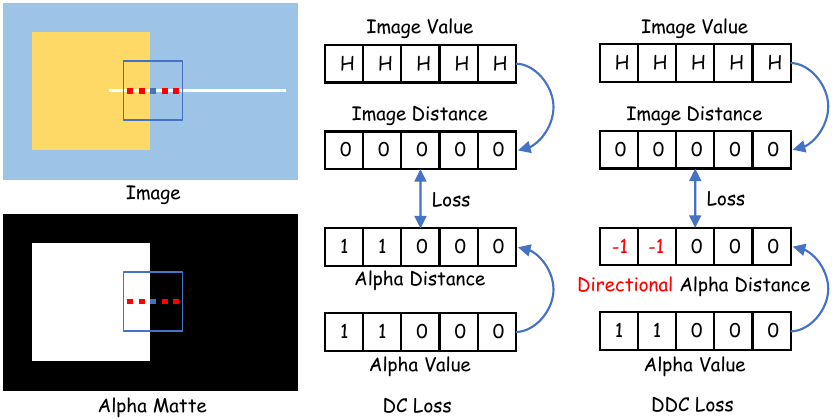}
	\caption{\textbf{A calculation instance of the proposed DC loss and DDC loss.} On a homogeneous hair whose pixel value is $H$, $K$ similar pixels are selected in the $K\times K$ window centered at a certain pixel. With the selected indices, the corresponding alpha values are gathered. DC loss first calculates the euclidean distances between the center pixel and the selected similar pixels both in the image and in the alpha matte, and then forces the two distance to be equal. Based on DC loss, DDC loss eliminates the interior noise by preserving the sign of alpha distance.
	}
	\label{fig:consistency}
\end{figure}

\subsection{Distance Consistency Loss}
\label{subsec:dcloss}

Based on the discussion above, we probe to add proper constraints to adapt nonlocal principle to the deep learning process. As discussed on Fig.~\ref{fig:braking}, affinity loss only stipulates linear or quadratic variation as a whole, but does not control the local trend. Therefore the affinity loss fails to provide enough punishment when alpha values tend to distribute as in the bottom right subplot. We address this by presenting a new assumption: if some pixels in the image are similar, then their corresponding alpha values are \textit{similar of the same degree}. Different from the affinity loss, we constrain pair-wise euclidean distance in the alpha matte and in the image to be equal. The new assumption brings about the distance consistency loss (DC Loss) shown in Fig.~\ref{eq:ddcloss}:
\begin{equation}
\begin{aligned}
    \mathcal{L}_\textrm{DC}=\frac{1}{N}\sum\limits_{i=1}^N\sum\limits_j\big|\Vert\alpha_i-\alpha_j\Vert_2-\Vert \boldsymbol{I}_i - \boldsymbol{I}_j\Vert_2\big|\,,\\
    j\in \textrm{argtopk}\{-\Vert \boldsymbol{I}_i - \boldsymbol{I}_j\Vert_2\}\,.
\end{aligned}
  \label{eq:dcloss}
\end{equation}
DC loss forces the variation of the alpha values to be the same as that in the image, so the smooth transition property at boundaries can be naturally satisfied. Moreover, DC loss governs the difference between local alpha pairs according to the pixel distance, which avoids the situation in the bottom right subplot in Fig.~\ref{fig:braking}. Fig.~\ref{fig:traing_alpha} (c) suggests the power of DC loss to fit subtle details.

\begin{figure}[!t]
	\centering
	\includegraphics[width=\linewidth]{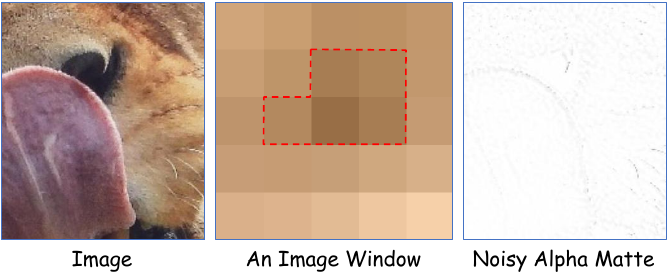}
	\caption{\textbf{Noise introduced by DC loss.} Owing to the body textures, the selected similar pixels (in the red dotted box) often have different values. DC loss forces the different values to exist also in the alpha matte. Such conflicts between DC loss and known loss introduces noise, appearing as undesired image textures in the alpha matte (zoom in the right part for best view).
	}
	\label{fig:conflict}
\end{figure}
\subsection{Directional Distance Consistency Loss}
\label{subsec:ddcloss}
Though DC loss helps a lot on subtle details, it causes conflicts against known loss in known regions. A random instance of noisy output is shown in Fig.~\ref{fig:conflict}. On the one hand, known loss forces `$1$' values on the foreground; on the other, DC loss expects certain variation because of the body textures. The loss confrontation introduces texture noise on the foreground/background. To make the two losses compatible, we update Eq.~\eqref{eq:dcloss} by preserving the sign of alpha distance as
\begin{equation}
\begin{aligned}
  \mathcal{L}_\textrm{DDC}=\frac{1}{N}\sum\limits_{i=1}^N\sum\limits_j\big|\alpha_i-\alpha_j-\Vert \boldsymbol{I}_i - \boldsymbol{I}_j\Vert_2\big|\,,\\
    j\in \textrm{argtopk}\{-\Vert \boldsymbol{I}_i - \boldsymbol{I}_j\Vert_2\}\,.
\end{aligned}
  \label{eq:ddcloss}
\end{equation}
The calculation process of DDC loss is shown on the right of Fig.~\ref{fig:consistency}, and the code can be found in the supplementary.
Consider a similar pair of pixel $i$ and $j$. If $j$ is selected as the similar pixel at the $i$-centered window, then $i$ will likely be selected at the window centered at $j$. Then DDC loss has a minimum of $2\Vert \boldsymbol{I}_i-\boldsymbol{I}_j\Vert$ at each similar pair, as deduced in the following absolute value inequality:
\begin{equation}
\begin{aligned}
    &\big|\alpha_i-\alpha_j-\Vert \boldsymbol{I}_i - \boldsymbol{I}_j\Vert\big|+\big|\alpha_j-\alpha_i-\Vert \boldsymbol{I}_i - \boldsymbol{I}_j\Vert\big| \\
    \geq&\big|\alpha_i-\alpha_j-\Vert \boldsymbol{I}_i - \boldsymbol{I}_j\Vert+\alpha_j-\alpha_i-\Vert \boldsymbol{I}_i - \boldsymbol{I}_j\Vert\big| \\
    =&2\Vert \boldsymbol{I}_i-\boldsymbol{I}_j\Vert\,.
\end{aligned}
\label{eq:lazy_neq}
\end{equation}
Note the condition of equality meets when $0\leq\vert\alpha_i-\alpha_j\vert\leq\Vert \boldsymbol{I}_i-\boldsymbol{I}_j\Vert$. When the pair are in the absolute foreground/background, known loss forces $\alpha_i=\alpha_j$, satisfying the condition of equality. At a cross section of the boundary, alpha values change monotonously. Let us just take $\alpha_i\geq\alpha_j$, then the condition of equality $0\leq\alpha_i-\alpha_j\leq\Vert \boldsymbol{I}_i-\boldsymbol{I}_j\Vert$ illustrates that the change rate of $\alpha$ is bounded by $\Vert \boldsymbol{I}_i-\boldsymbol{I}_j\Vert$, so the smooth transition property still holds. In short, we set a positive minimum for DC loss, where the know loss also reaches its minimum $0$.

\vspace{5pt}
\noindent\textbf{Total loss.} The total loss combines known loss and DDC loss as
\begin{equation}
  \mathcal{L}_\textrm{total}=\mathcal{L}_\textrm{known}+\lambda\mathcal{L}_\textrm{DDC}\,,
  \label{eq:total_loss}
\end{equation}
where $\lambda$ is a positive number to control the impact of the two types of losses. Note that $\mathcal{L}_\textrm{known}$ only supervises regions of absolute foreground/background indicated by trimaps with $1$ and $0$, and $\mathcal{L}_\textrm{DDC}$ supervises the whole alpha matte.

%% file: sec/4_experiment.tex
\section{Experiment}
\label{sec:experiment}
\begin{figure*}[!t]
	\centering
	\includegraphics[width=\linewidth]{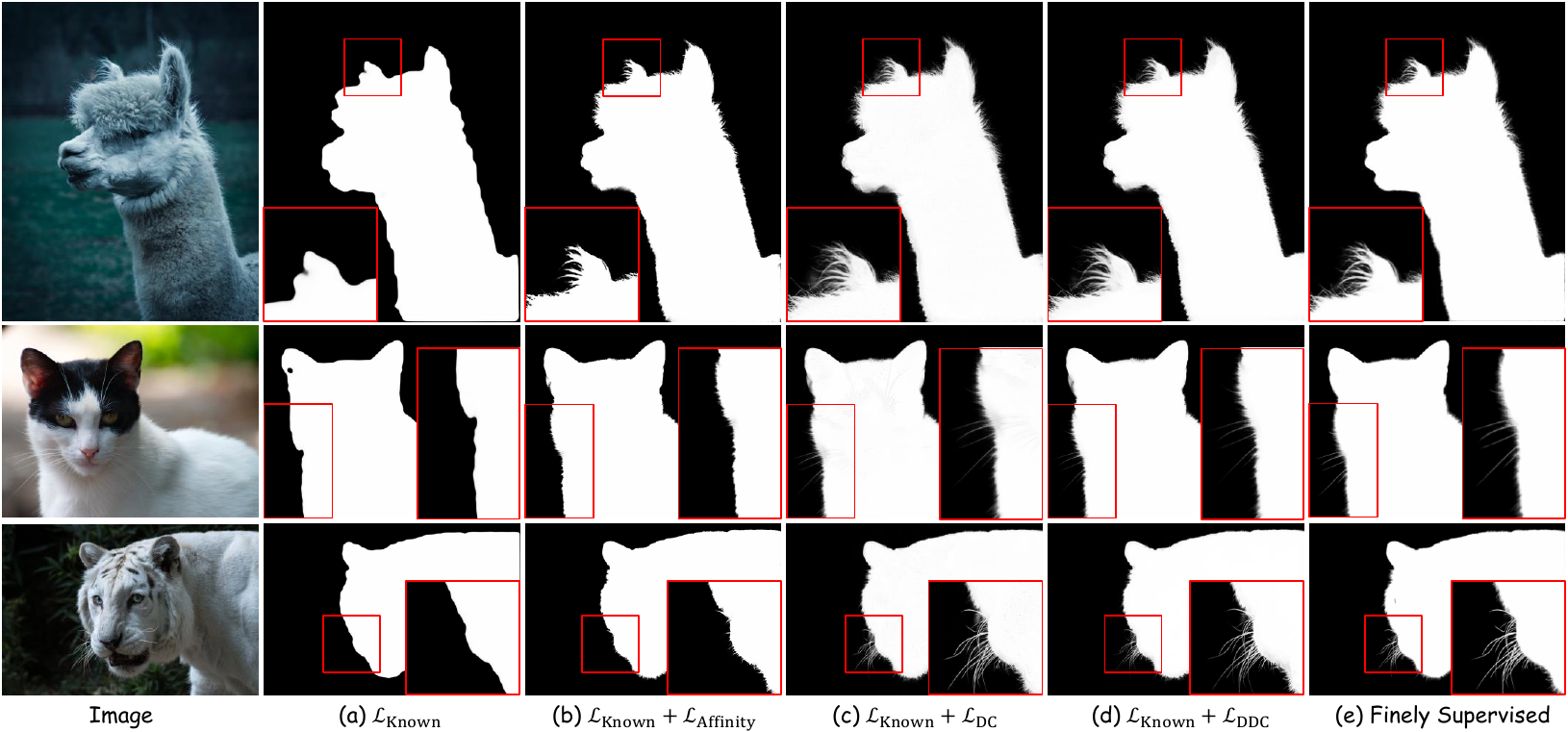}
	\caption{\textbf{Visual results of trimap-supervised baselines described in Sec.~\ref{sec:method} and the finely supervised baseline.} The examples are chosen from the AM-2K~\cite{li2022bridging} test set and the P3M-NP-500 test set of P3M-10K~\cite{li2021privacy}. (a)-(d) For the four trimap-supervised baselines, the inference results coincide with the study in Sec.~\ref{sec:method}. (e) Our trained model predicts similar alpha mattes with the finely supervised baseline.
	}
	\label{fig:policy_visualization}
\end{figure*}
In this section, we verify the effectiveness of our training paradigm. Automatic matting is chosen as an application example for its simplicity. The paradigm can similarly be applied on other matting modes.
\subsection{Implementation Details}

\vspace{5pt}
\noindent\textbf{Network architecture.} The proposed paradigm can be applied to any deep matting model. For convenience, we choose the recent ViTMatte~\cite{yao2023vitmatte} model, which is composed of a plain ViT~\cite{dosovitskiy2020image} backbone and a lightweight decoder. 

\vspace{5pt}
\noindent\textbf{Evaluation.} We report the common used metrics of Sum of Absolute Differences (SAD), Mean Squared Error (MSE), Gradient (Grad) and Connectivity (Conn) proposed by~\cite{rhemann2009perceptually} and Mean Absolute Difference (MAD) following~\cite{li2022bridging} to evaluate the quantitative performance. Among the metrics, SAD, MAD and MSE compare the pixel differences between the prediction results and the ground truth alpha matte. Grad evaluates fine details such as hairs, which increases when encountering oversmoothing predictions. Conn tests the connectivity of the foreground/background, which grows with artifacts or holes in the prediction. We first evaluate the metrics with all pixels, and then test the SAD and MSE metric at transition areas indicated by $0.5$ in the trimap (SAD-T and MSE-T).

\vspace{5pt}
\noindent\textbf{Dataset.} 
Affected by domain gap, models trained on synthetic data (Composition-1K~\cite{xu2017deep}, Distinct 646~\cite{qiao2020attention}, etc.) often work poorer in reality. Without requiring fine labels, the proposed method does not rely on data synthesis to produce labels. Hence, we verify the effectiveness of our method directly on natural datasets AM-2K~\cite{li2022bridging} and P3M-10K~\cite{li2021privacy}.
The AM-2K dataset is an animal matting dataset, with a total of $2000$ image samples of multiple animal categories. Among all the image/label pairs, $1800$ samples are used for training and the rest $200$ are left for validation. The P3M-10K dataset~\cite{li2021privacy} is used for automatic portrait matting, which consists of $10421$ face-blurred portrait images. We use its $9421$-sample training set, and choose one of its validation set P3M-500-NP for testing. The model is separately trained and evaluated on the two datasets. We use alpha erosion~\cite{xu2017deep} with the erosion kernel size randomly chosen in $[1,30]$ to produce trimaps as stimulation of human labelling. The study on the effect of trimap roughness can be found in the supplementary. During training only generated trimaps are used as labels.

\vspace{5pt}
\noindent\textbf{Training setting.} In data augmentation, we apply the common used random scale with the scaling factor between $0.5$ and $2$, random crop to size of $512\times512$, and horizontal flip with a probability of $0.5$ onto the image and its corresponding trimap for AM-2K dataset~\cite{li2022bridging}. And we use solely random crop to $512\times512$ for P3M-10K dataset~\cite{li2021privacy}. For the model architecture, we use ViTMatte-S~\cite{yao2023vitmatte} with the ViT-S~\cite{dosovitskiy2020image} backbone pretrained with DINO~\cite{caron2021emerging} by default. The AdamW optimizer with the initial learning rate of $5e-4$ and the weight decay rate of $0.1$ is adopted to train the model. The training period is set as $100$ epochs with the learning rate multiplying $0.1$ and $0.05$ at the $60$-th and $90$-th epoch respectively for animal matting, while in portrait matting the total number of epochs is $500$, and the learning rate decays at the $300$-th and $450$-th epoch. Following ViTMatte~\cite{yao2023vitmatte}, a layerwise learning rate is applied with a delay rate of $0.65$ during fine-tuning. We use $4$ NVIDIA GeForce RTX 3090 GPUs with a total batch size of $16$ for training. In the expression Eq.~\eqref{eq:ddcloss} of DDC loss, the window size $K$ is set as $11$ by default. For the total loss Eq.~\eqref{eq:total_loss}, $\lambda$ is set as $10$ to balance the impact of the two losses. Following ViTMatte~\cite{yao2023vitmatte}, $\mathcal{L}_\textrm{known}$ is 
additionally multiplied by a scaling factor $s=\frac{N}{N_\textrm{known}}$, where $N$ and $N_\textrm{known}$ is the total pixel number and the known pixel number respectively.
\subsection{Main Results}

\vspace{5pt}
\noindent\textbf{Validation of the effectiveness.} The effectiveness of our proposed paradigm should be assessed by comparison against the fine-label-supervised counterpart. Thus we build two baselines as illustrated in Fig.~\ref{fig:paradigm} (a). The first one is supervised using fine alpha mattes with an $l_1$ loss, and the second one adopts only trimaps as labels under supervision of known loss (an $l_1$ loss) and the proposed DDC loss. Other settings keep the same. The quantitative results of the two baselines correspond to Row 1 and Row 2 in Table~\ref{tab:ablation_label} respectively. It demonstrates that our paradigm has comparable performance with the finely supervised baseline, and even better on the Grad and MSE metric. The visual comparison can be referred to Fig.~\ref{fig:policy_visualization} (d) and (e). As one can see, the model trained under our paradigm by only coarse trimap labels predicts similar alpha mattes with the counterpart trained by fine alpha matte annotations.

\begin{table}\scriptsize
  \centering
  \addtolength{\tabcolsep}{-1pt}
  \begin{tabular}{@{}lccccccccc@{}}
    \toprule
    Label & SAD & MAD & MSE & Grad & Conn & SAD-T & MSE-T \\
    \midrule
    Matte & 26.00 & 0.0149 & 0.0101 & 16.04 & 10.05 & 10.82 & 0.0268 \\
    \rowcolor{lightgray!30} Trimap & 30.26 & 0.0175 & \textbf{0.0099} & \textbf{14.98} & 14.33 & 16.07 & 0.0356 \\
    \bottomrule
  \end{tabular}
  \caption{Comparison on the training paradigm.}
  \label{tab:ablation_label}
\end{table}

\begin{figure*}[!t]
	\centering
	\includegraphics[width=\linewidth]{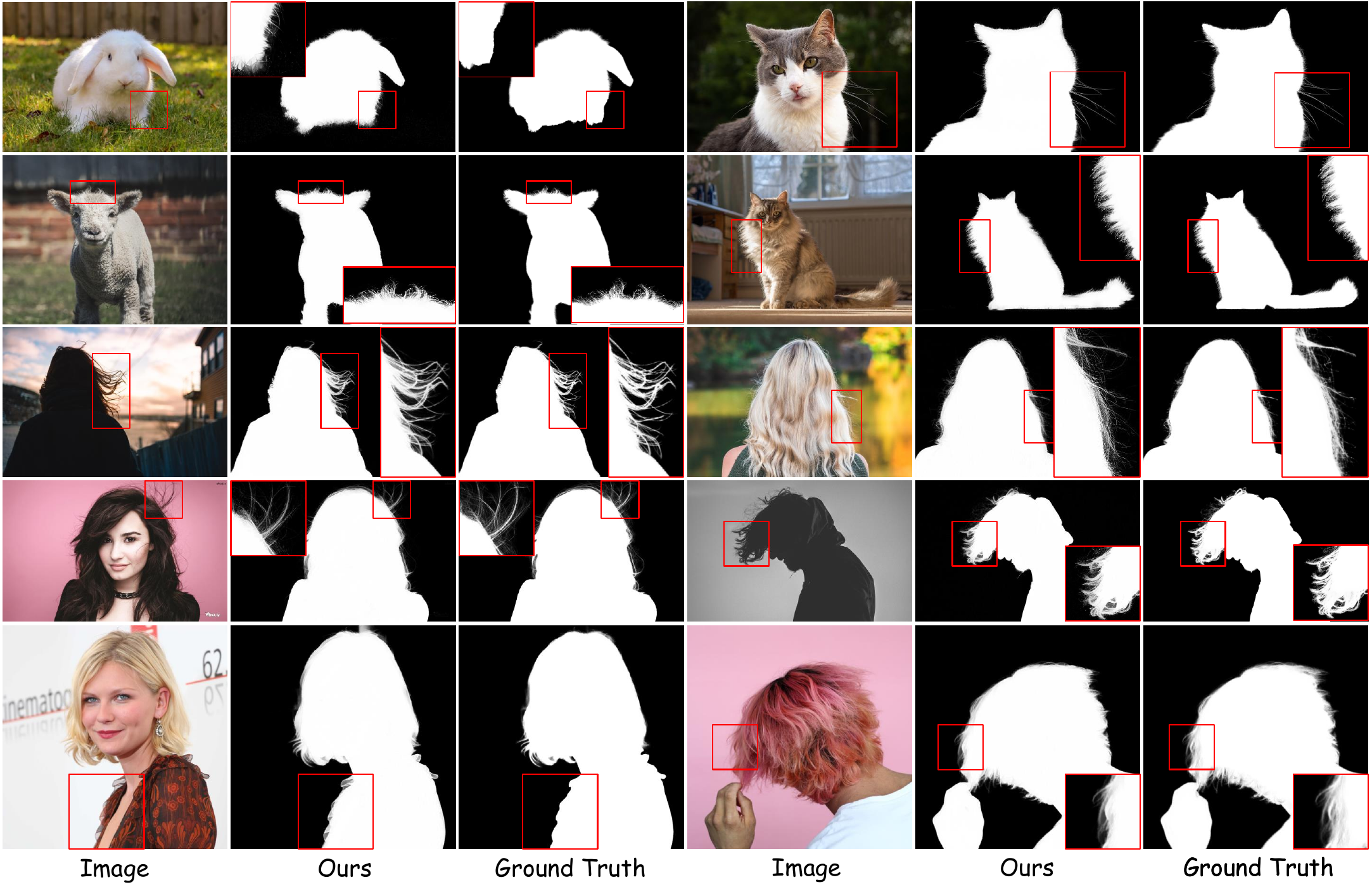}
	\caption{\textbf{Visualization samples on the AM-2K~\cite{li2022bridging} test set and the P3M-NP-500 test set of P3M-10K~\cite{li2021privacy}.}
	}
	\label{fig:prediction_visual}
\end{figure*}

\vspace{5pt}
\noindent\textbf{Image matting results.} We compare our trained model with several recent automatic matting methods. The automatic animal matting results on the AM-2K~\cite{li2022bridging} test set and the automatic portrait matting results on P3M-NP-500 test set of P3M-10K~\cite{li2021privacy} are shown in Table~\ref{tab:animal} and Table~\ref{tab:portrait} respectively. According to the results, our trained model obtains comparable performance with recent automatic matting approaches. We provide several examples in Fig.~\ref{fig:prediction_visual}. One can see that our trained model predicts subtle details for both animals and humans. Further more, because our model is directly trained under matting priors, it can sometimes produce results more correct than the human-labelled ground truth. For example, in the left subfigure of Row 1, ours produces more detailed alpha values on the conjunction of the rabbit and the grass; in the left subfigure of the last row, ours well delineate the chiffon on the clothes, while it is considered as opaque in the ground truth.

\begin{table}\scriptsize
  \centering
  \addtolength{\tabcolsep}{-3pt}
  \begin{tabular}{@{}lcccccccc@{}}
    \toprule
    Method & Label & SAD & MAD &  MSE & Grad & Conn & SAD-T & MSE-T \\
    \midrule
    SHM~\cite{chen2018semantic} & Matte & 17.81 & 0.0102 & 0.0068 & 12.54 & 17.02 & 10.26 & - \\
    LF~\cite{zhang2019late} & Matte & 36.12 & 0.0210 & 0.0116 & 21.06 & 33.62 & 19.68 & - \\
    SSS~\cite{aksoy2018semantic} & Matte & 552.88 & 0.3225 & 0.2742 & 60.81 & 555.97 & 88.23 & - \\
    HATT~\cite{qiao2020attention} & Matte & 28.01 & 0.0161 & 0.0055 & 18.29 & 17.76 & 13.36 & - \\
    GFM~\cite{li2022bridging} & Matte & 10.26 & 0.0059 & 0.0029 & 8.82 & 9.57 & 8.24 & - \\
    \rowcolor{lightgray!30} Ours & Trimap & 30.26 & 0.0175 & 0.0099 & 14.98 & 14.33 & 16.07 & 0.0356 \\
    \bottomrule
  \end{tabular}
  \caption{Automatic animal matting results on AM-2K~\cite{li2022bridging} test set.}
  \label{tab:animal}
\end{table}
\begin{table}\scriptsize
  \centering
  \addtolength{\tabcolsep}{-3pt}
  \begin{tabular}{@{}lcccccccccc@{}}
    \toprule
    Method & Label & SAD & MAD & MSE & Grad & Conn & SAD-T & MSE-T\\
    \midrule
    LF~\cite{zhang2019late} & Matte & 32.59 & 0.0188 & 0.0131 & 31.93 & 19.50 & 14.53 & 0.0420 \\
    HATT~\cite{qiao2020attention} & Matte & 30.53 & 0.0176 & 0.0072 & 19.88 & 27.42 & 13.48 & 0.0403 \\
    SHM~\cite{chen2018semantic} & Matte & 20.77 & 0.0122 & 0.0093 & 20.30 & 17.09 & 9.14 & 0.0255 \\
    MODNet~\cite{ke2022modnet} & Matte & 16.70 & 0.0097 & 0.0051 & 15.29 & 13.81 & 9.13 & 0.0237 \\
    GFM~\cite{li2022bridging} & Matte & 15.50 & 0.0091 & 0.0056 & 14.82 & 18.03 & 10.16 & 0.0268 \\
    P3M~\cite{li2021privacy} & Matte & 11.23 & 0.0065 & 0.0035 & 10.35 & 12.51 & 5.32 & 0.0094 \\
    \rowcolor{lightgray!30} Ours & Trimap & 31.66 & 0.0182 & 0.0126 & 15.41 & 13.66 & 12.03 & 0.0371 \\
    \bottomrule
  \end{tabular}
  \caption{Automatic portrait matting results on P3M-NP-500~\cite{li2021privacy}.}
  \label{tab:portrait}
\end{table}

\subsection{Ablation Study}
\begin{figure}[!t]
	\centering
	\includegraphics[width=\linewidth]{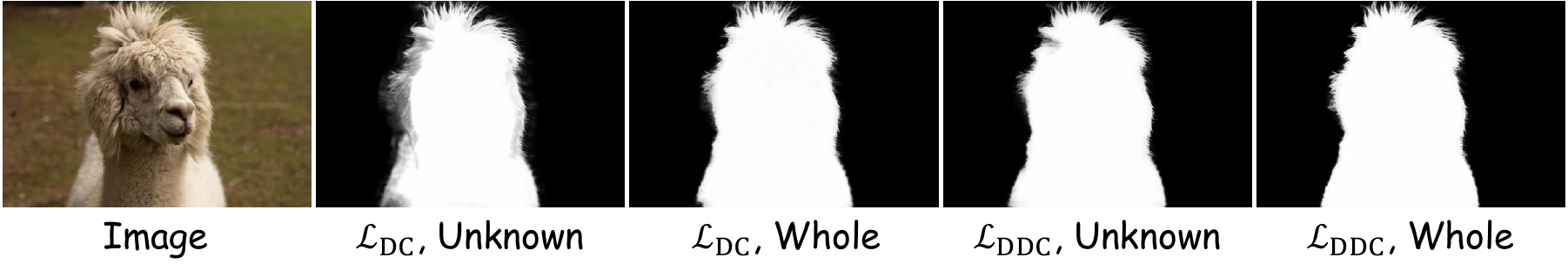}
	\caption{\textbf{Visual comparison of whether supervising the whole alpha map. }
	}
	\label{fig:supervise_region}
\end{figure}
\vspace{5pt}
\noindent\textbf{Different loss policies.} Here we compare the four loss policies mentioned in Sec.~\ref{sec:method}. Row 1-4 of Table~\ref{tab:ablation_policy} provides the quantitative performance of each loss policy corresponds to the visualizations in Fig.~\ref{fig:traing_alpha} (a)-(d) respectively. While all supervised by trimaps, Row 1 uses only the known loss, Row 2 adds affinity loss described by Eq.~\eqref{eq:affinity_loss} as assistance of the known loss, Row 3 instead adopts known loss and DC loss in Eq.~\eqref{eq:dcloss}, and Row 4 replaces DC loss with DDC loss based on Row 3. Fig.~\ref{fig:traing_alpha} shows that known loss can only provide a coarse mask, in accord with the large SAD error in Table~\ref{tab:ablation_policy}. Affinity loss can help fit a few details and reduce the errors. Compared with affinity loss, DC loss invites more subtle details and largely reduces the Grad error by more than $7$ points. The visual effect can be verified by the cat beard in the red box. However, due to the texture noise introduced (Fig.~\ref{fig:traing_alpha} (c) and Fig.~\ref{fig:conflict}), DC loss even increases the SAD error. The updated version DDC loss fits the details well and preserves the constant values in known areas, contributing high-quality alpha matte predictions. Quantitatively, DDC loss reduces the Grad error by $8.43$ points while remains other metrics comparable with affinity loss.
\begin{table}\scriptsize
  \centering
  \addtolength{\tabcolsep}{-2.5pt}
  \begin{tabular}{@{}cccccccc@{}}
    \toprule
    Loss policy & SAD & MAD & MSE & Grad & Conn & SAD-T & MSE-T \\
    \midrule
    $\mathcal{L}_\textrm{known}$ & 45.94 & 0.0266 & 0.0203 & 77.70 & 12.35 & 29.89 & 0.1096 \\
    $\mathcal{L}_\textrm{known}+\mathcal{L}_\textrm{affinity}$ & 32.54 & 0.0188 & 0.0132 & 23.41 & \textbf{11.00} & 17.25 & 0.0496 \\
    $\mathcal{L}_\textrm{known}+\mathcal{L}_\textrm{DC}$ & 42.24 & 0.0247 & 0.0117 & 16.07 & 21.59 & 18.82 & 0.0396 \\
    \rowcolor{lightgray!30} $\mathcal{L}_\textrm{known}+\mathcal{L}_\textrm{DDC}$ & \textbf{30.26} & \textbf{0.0175} & \textbf{0.0099} & \textbf{14.98} & 14.33 & \textbf{16.07} & \textbf{0.0356} \\
    \bottomrule
  \end{tabular}
  \caption{Comparison among four training policies.}
  \label{tab:ablation_policy}
\end{table}

\vspace{5pt}
\noindent\textbf{The effect of supervision regions.} It seems that the conflict introduced by DC loss vanishes if DC loss only supervises unknown areas. In fact, if doing so, the unknown values would manifest large discontinuity against known regions as shown in Fig.~\ref{fig:supervise_region}. The reason behind is that neither known loss nor DC loss supervises the unknown values, and DC loss only supervises the distance relation. As a result, the unknown values would adaptively change to decrease DC loss, while not consider the consistency with known regions. In short, missing constraints on known regions by DC or DDC loss hurts the smooth transition property, which is also quantitatively verified by Table~\ref{tab:ablation_region}. The effect of supervision regions also implies that coarse trimaps are better than binary segmentation labels, which can be further referred to the supplementary.

\begin{table}\scriptsize
  \centering
  \addtolength{\tabcolsep}{-2.5pt}
  \begin{tabular}{@{}ccccccccc@{}}
    \toprule
    Loss & Region & SAD & MAD & MSE & Grad & Conn & SAD-T & MSE-T \\
    \midrule
    $\mathcal{L}_\textrm{DC}$ & Unknown & 72.62 & 0.0422 & 0.0204 & 30.06 & 48.34 & 37.02 & 0.0828 \\
    $\mathcal{L}_\textrm{DC}$ & Whole & 42.24 & 0.0247 & 0.0117 & 16.07 & 21.59 & 18.82 & 0.0396 \\
    $\mathcal{L}_\textrm{DDC}$ & Unknown & 38.80 & 0.0225 & 0.0141 & 18.75 & 18.34 & 20.27 & 0.0514 \\
    \rowcolor{lightgray!30} $\mathcal{L}_\textrm{DDC}$ & Whole & \textbf{30.26} & \textbf{0.0175} & \textbf{0.0099} & \textbf{14.98} & \textbf{14.33} & \textbf{16.07} & \textbf{0.0356} \\
    \bottomrule
  \end{tabular}
  \caption{Ablation study on whether supervising the whole alpha map by DC or DDC loss. Each is with known loss.}
  \label{tab:ablation_region}
\end{table}

\vspace{5pt}
\noindent\textbf{The effect of window size.} Large window is required to make sure that the selected pixels are similar in color to trace long-range dependency. The experimental results on kernel size can be found in the supplementary materials. We empirically find that $K=11$ is the best design choice. Too small kernel struggle to model long-range dependency, while too large ones easily introduces noise.

%% file: sec/5_conclusion.tex
\section{Conclusion}
In this paper, we present a coarsely supervised training paradigm to rid deep image matting of fine labels. In the proposed paradigm, we use coarse trimap labels to enable the model to distinguish the foreground/background, and devise a novel directional distance consistency loss (DDC loss) to control the alpha values at transition areas conditioned on the image. DDC loss forces the local distance between similar neighbors on the alpha matte and on the corresponding image to be equal. Experiments prove that the proposed paradigm yields high-quality matting predictions comparable to the finely supervised baseline. For future work, we will explore transparent objects matting and other image matting tasks such as interactive matting.

%% file: sec/X_suppl.tex
\clearpage

\section{Appendix}


\renewcommand\thesubsection{\Alph{subsection}}
\subsection{Impact of Trimap Roughness}
\label{subsec:trimap_roughness}
\begin{figure}[!t]
	\centering
	\includegraphics[width=\linewidth]{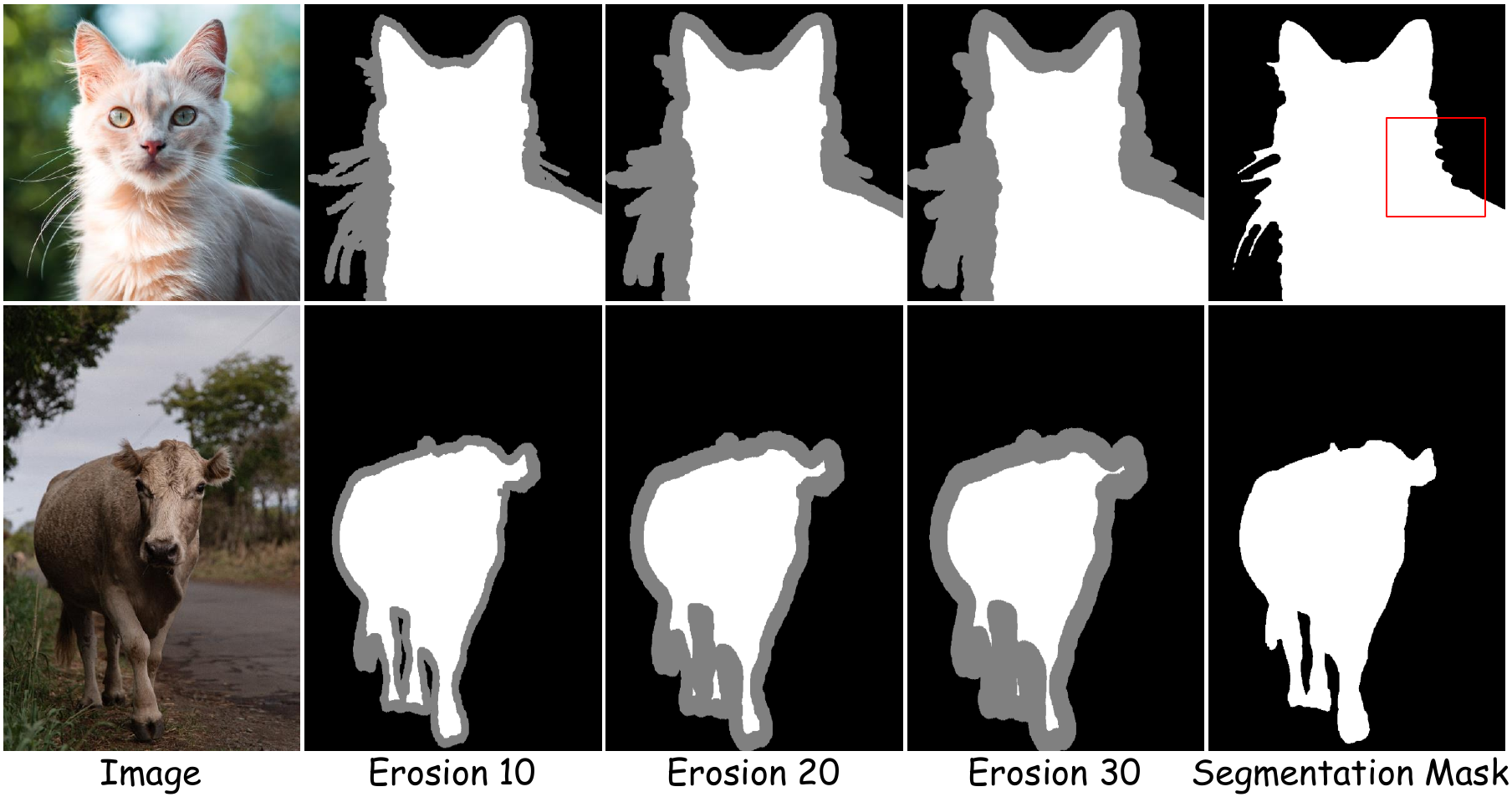}
	\caption{\textbf{Trimaps generated by different erosion kernel size and segmentation masks.} Large erosion kernel may affect the perception of object shapes, \eg, the bull legs. Segmentation masks forces wrong predictions, for example in the red box some hairs are recognized as background (black color).
	}
	\label{fig:erosion}
\end{figure}

To simulate deviations in human labeling habits, we use a random erosion kernel size in $[1,30]$ by default in Sec.~\ref{sec:experiment}. Here we test the effect of trimap roughness by setting different erosion kernel sizes including $10$, $20$ and $30$ to generate trimap labels. The quantitative results are shown in Table~\ref{tab:ablation_stability}, where finer trimap annotations invite better performance. An explanation is that rougher trimaps are more adverse to the perception of object shapes. As one can see in the lower part of Fig.~\ref{fig:erosion}, large erosion kernels can impair the description of the bull legs.

\subsection{Training with Segmentation Mask Labels}
\label{subsec:mask_supervision}

The coarse segmentation mask is another choice of coarse labels. However, such masks force wrong predictions because of full supervision, which hinders the model to fit subtle details during training. For an instance as shown in the red box of Fig.~\ref{fig:erosion}, some hairs are recognized as black-colored background. The quantitative results are shown in Table~\ref{tab:ablation_mask}, where two types of known losses including $l_1$ loss and binary cross entropy loss (BCE loss) are tested. One can see that training with trimap labels outperforms the counterpart with segmentation mask labels by a large margin. As a result, our paradigm chooses trimaps as the coarse labels.

\subsection{Effect of $d_{ij}^2$ in Affinity Loss}
\label{subsec:d_ij}

\begin{figure}[!t]
	\centering
	\includegraphics[width=\linewidth]{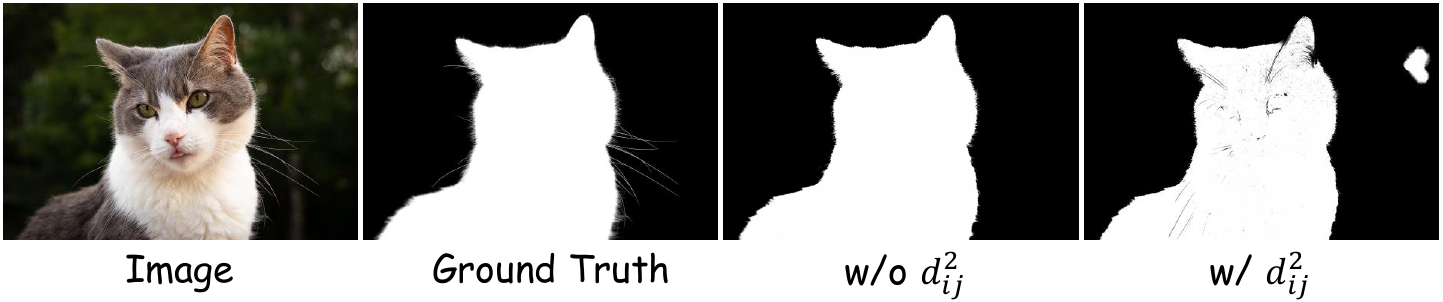}
	\caption{\textbf{Effect of the item $d_{ij}^2$ in affinity loss.} $d_{ij}^2$ introduces noise into the calculation of pixel distance.
	}
	\label{fig:spatial_distance}
\end{figure}

In Sec.~\ref{subsec:preliminary} we claim that the item $d_{ij}^2$ introduces noise into the selection of similar pixels. Here we provide visual supports in Fig.~\ref{fig:spatial_distance}. Between the two predictions, the left is trained with $f_1(i,j)=\Vert \boldsymbol{I}_i - \boldsymbol{I}_j\Vert_2$ and the right is trained with $f_2(i,j)=\Vert \boldsymbol{I}_i - \boldsymbol{I}_j\Vert_2 + d_{ij}^2$. The similar pixels are chosen in each $K\times K$ window with $f_1$, while they are selected globally with $f_2$ under the help of the FRNN~\footnote{https://github.com/lxxue/FRNN} algorithm for approximate nearest neighbor search. Fig.~\ref{fig:spatial_distance} manifests strong noise introduced by the item $d_{ij}^2$.




\subsection{Effect of Window Size in DDC Loss}
\label{subsec:window_ddc}

We report quantitative ablation results on the kernel size $K$ involved in DDC loss in Table~\ref{tab:ablation_window}, where the kernel sizes range from $5$ to $13$. It suggests that $K=11$ achieves the best trade-off between the workload and the performance for DDC loss.

\begin{table}\scriptsize
  \centering
  \addtolength{\tabcolsep}{-1.5pt}
  \begin{tabular}{@{}cccccccccc@{}}
    \toprule
    Kernel Size & SAD & MAD & MSE & Grad & Conn & SAD-T & MSE-T\\
    \midrule
    10 & 31.78 & 0.0183 & 0.0110 & 15.76 & 13.53 & 15.82 & 0.0370 \\
    20 & 42.80 & 0.0250 & 0.0155 & 17.13 & 18.82 & 21.05 & 0.0503 \\
    30 & 44.48 & 0.0257 & 0.0140 & 18.72 & 24.16 & 26.13 & 0.0628 \\
    \rowcolor{lightgray!30} Rand(1,30) & 30.26 & 0.0175 & 0.0099 & 14.98 & 14.33 & 16.07 & 0.0356\\
    \bottomrule
  \end{tabular}
  \caption{Study on erosion kernel size for trimap production.}
  \label{tab:ablation_stability}
\end{table}

\begin{table}\scriptsize
  \centering
  \addtolength{\tabcolsep}{-3pt}
  \begin{tabular}{@{}lcccccccccc@{}}
    \toprule
    Label & Known loss & SAD & MAD & MSE & Grad & Conn & SAD-T & MSE-T\\
    \midrule
    Mask & $l_1$ & 40.47 & 0.0233 & 0.0155 & 23.38 & 15.61 & 22.06 & 0.0608 \\
    Mask & BCE & 61.03 & 0.0352 & 0.0136 & 24.18 & 32.05 & 35.46 & 0.0643 \\
    \rowcolor{lightgray!30} Trimap & $l_1$ & 30.26 & 0.0175 & 0.0099 & 14.98 & 14.33 & 16.07 & 0.0356 \\
    \bottomrule
  \end{tabular}
  \caption{Comparison between using trimap labels and using segmentation mask labels. All losses are with the proposed DDC loss.}
  \label{tab:ablation_mask}
\end{table}

\begin{figure}[!t]
	\centering
	\includegraphics[width=\linewidth]{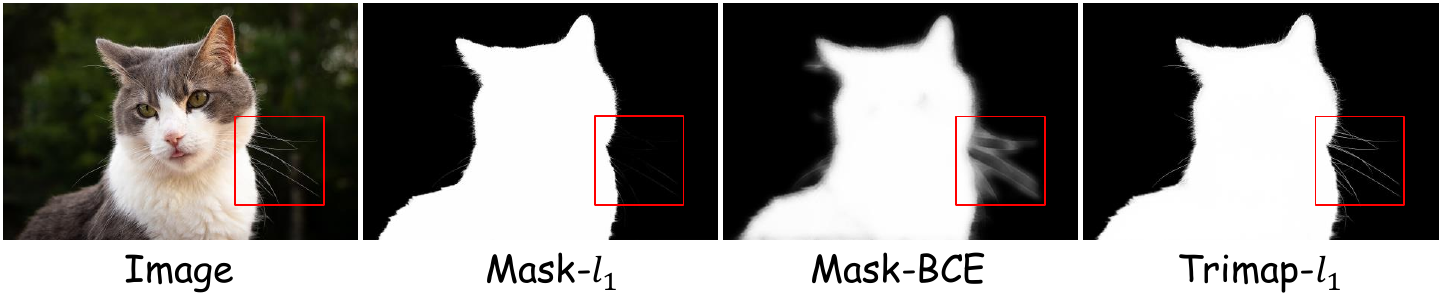}
	\caption{\textbf{Visual comparison between segmentation masks and trimaps as coarse labels.} Segmentation mask labels with $l_1$ loss suppresses the prediction of details, while that with BCE loss tends to generate vague predictions. On the contrary, Trimap labels with $l_1$ loss predicts high-quality alpha mattes.
	}
	\label{fig:coarse_label_type}
\end{figure}

\begin{table}\scriptsize
  \centering
  \addtolength{\tabcolsep}{-1.5pt}
  \begin{tabular}{@{}cccccccccc@{}}
    \toprule
    Window Size & SAD & MAD & MSE & Grad & Conn & SAD-T & MSE-T\\
    \midrule
    5 & 41.01 & 0.0241 & 0.0154 & 17.78 & 17.21 & 19.93 & 0.0473 \\
    7 & 37.10 & 0.0214 & 0.0127 & 16.04 & 17.42 & 19.28 & 0.0433 \\
    9 & 33.91 & 0.0197 & 0.0114 & 15.48 & 16.48 & 17.58 & 0.0400 \\
    \rowcolor{lightgray!30} 11 & \textbf{30.26} & \textbf{0.0175} & \textbf{0.0099} & \textbf{14.98} & 14.33 & 16.07 & \textbf{0.0356} \\
    13 & 33.01 & 0.0189 & 0.0115 & 16.54 & 14.26 & 16.68 & 0.0402 \\
    15 & 35.61 & 0.0203 & 0.0130 & 16.59 & \textbf{14.14} & \textbf{15.88} & 0.0372 \\
    \bottomrule
  \end{tabular}
  \caption{Ablation study on window size for similar pixel selection.}
  \label{tab:ablation_window}
\end{table}

\subsection{Sample Code for DDC Loss}
\label{subsec:code_ddc}
A sample code is provided in Listing~\ref{listing:code_ddc}. With only several lines of code added, any image matting model can be trained with trimap labels. The code is based on PyTorch.
\lstset{
    language=Python,
    basicstyle=\ttfamily\footnotesize,
    tabsize=4,
    numbers=left,
    numberstyle=\tiny\color{gray},
    breaklines=true,
    captionpos=b,
    keywordstyle=\color{magenta},
    commentstyle=\color{LimeGreen},
}
\begin{lstlisting}[caption=Sample code for DDC loss., label=listing:code_ddc,]
import torch
import torch.nn.functional as F

def ddc_loss(image, alpha, kernel_size):
    # each value in image is in [0,1]
    b, c, h, w = image.shape
    unfold_image = F.unfold(image, kernel_size=kernel_size, padding=kernel_size // 2).view(b, c, kernel_size ** 2, h, w)
    image_dist = torch.norm(image.view(b, c, 1, h, w) - unfold_image, 2, dim=1)
    image_dist, indices = torch.topk(image_dist, k=kernel_size, dim=1, largest=False)
    unfold_alpha = F.unfold(alpha, kernel_size=kernel_size, padding=kernel_size // 2).view(b, kernel_size ** 2, h, w)
    alpha_dist = torch.gather(alpha - unfold_alpha, dim=1, index=indices)
    return F.l1_loss(image_dist, alpha_dist)
\end{lstlisting}